\documentclass[10pt]{article}

\usepackage[utf8]{inputenc}
\usepackage[T1]{fontenc}
\usepackage{microtype}
\usepackage{graphicx}
\usepackage[margin=1in]{geometry}
\usepackage{authblk}
\usepackage{url}
\usepackage{booktabs}
\usepackage{amsfonts}
\usepackage{amsmath}
\usepackage{amssymb}
\usepackage{nicefrac}
\usepackage{lipsum}
\usepackage{caption}
\usepackage[backend=biber]{biblatex}
\usepackage{subcaption}
\usepackage{hyperref}
\usepackage{float}
\usepackage{geometry}
\usepackage{tabularx} % Allows better column width control
\usepackage[table]{xcolor}
\definecolor{lightgray}{RGB}{235,235,235}

\newcommand{\modelname}{\textsc{ORI}}

\title{\modelname: O Routing Intelligence\\
\large Vector Space-Driven Query Routing in Heterogeneous Language Model System}

\author[1]{Ahmad Shadid}
\author[1]{Rahul Kumar}
\author[1]{Mohit Mayank}

\affil[1]{O.SYSTEMS Foundation}

\date{}

\begin{document}

\maketitle
\begin{abstract}
Single large language models (LLMs) often fall short when faced with the ever-growing range of tasks, making a single-model approach insufficient. We address this challenge by proposing ORI (O Routing Intelligence), a dynamic framework that leverages a set of LLMs. By intelligently routing incoming queries to the most suitable model, ORI not only improves task-specific accuracy, but also maintains efficiency. Comprehensive evaluations across diverse benchmarks demonstrate consistent accuracy gains while controlling computational overhead. By intelligently routing queries, ORI outperforms the strongest individual models by up to 2.7 points on MMLU and 1.8 points on MuSR, ties the top performance on ARC, and on BBH. These results underscore the benefits of a multi-model strategy and demonstrate how ORI’s adaptive architecture can more effectively handle diverse tasks, offering a scalable, high-performance solution for a system of multiple large language models.
\end{abstract}
\section{Introduction}
The rapid advancement and increasing complexity of large language models (LLMs) have revolutionized the capabilities of natural language processing systems, allowing them to undertake a range of queries from simple classification to deep semantic analysis and interactive communication [1][2]. The introduction of the Transformer architecture, which is entirely based on attention mechanisms, has been pivotal in this evolution, enabling significant improvements in both performance and scalability [3].
Despite the transformative potential of these models, practical application remains constrained by significant computational costs and performance variability [4]. Recent efforts in model optimization focus on balancing these costs with performance, ensuring efficient deployment across various domains.

A critical advancement in this field has been the development of routing strategies that dynamically allocate queries to the most suitable models.
The MetaLLM [5] framework dynamically routes each query to the optimal LLM among several available models, achieving improved accuracy and cost-effectiveness. The OptLLM [6] framework addresses the cost-effective query allocation problem by predicting the performance of candidate LLMs on each query and providing users with a range of optimal solutions that align with their budget constraints and performance preferences. Similarly, RouterDC [7] employs a query-based router trained with dual contrastive learning to select the most suitable LLM for each query, effectively assembling multiple standard LLMs and outperforming individual top-performing models as well as existing routing methods.

Techniques like RouterDC [7] and RouteLLM [8] illustrate the potential of sophisticated routing mechanisms that integrate dual contrastive learning and preference data to better match queries with appropriate LLMs, outperforming traditional ensemble methods in both efficiency and efficacy.

Moreover, recent research investigates LLM routing for challenging reasoning tasks, highlighting the need for robust methods to harness the capabilities of different LLMs more efficiently [8]. These models use a variety of criteria, including task difficulty and model proficiency, to intelligently distribute computational load and optimize the balance between cost and performance.

The introduction of frameworks such as AutoMix [9], which dynamically mixes different model capabilities, represents another significant step toward a cost-effective deployment of LLM. This framework uses real-time decision making to allocate computational resources, which is crucial for applications that require high performance and fiscal prudence. Similarly, the development of the RouteLLM [8] framework, which incorporates data on human preference and data augmentation to refine routing decisions, exemplifies the potential of machine learning models to effectively reduce costs by more than twofold without compromising the quality of the output.

These developments underscore a broader move towards more sustainable and economically viable NLP solutions. By integrating these advanced methodologies, the field is poised to overcome current limitations, paving the way for wider adoption and more innovative applications of LLM technologies.
However, existing routing frameworks still face challenges in maintaining consistent performance on diverse benchmarks and efficiently managing computational resources. Some frameworks heavily depend on human preference data and data augmentation techniques to train their routers, which can introduce biases and may not generalize well across diverse tasks and domains. Additionally, certain models lack the flexibility to handle varying query complexities, leading to suboptimal performance when faced with diverse or unforeseen tasks. Moreover, some approaches involve the assembly of multiple LLMs or the use of complex routing mechanisms, resulting in increased computational overhead and latency. Furthermore, the effectiveness of these frameworks can vary across different benchmarks, indicating inconsistent performance and the need for more robust evaluation methods.

In this paper , our key contributions are as follows:
\begin{enumerate}
    \item \textbf{Granular Task Identification in Overlapping Benchmarks} \\
    We delve beyond top-level benchmarks to isolate more granular task structures, recognizing that a model that performs well on an entire benchmark may falter on specific sub-tasks. Clustering analyzes provide evidence for these nuanced task differences, ensuring more accurate and context-sensitive model selection.
    
    \item \textbf{Dimension-Flexible Optimization with ORI} \\
The ORI framework unifies multiple models and empirically determines which one excels under given constraints such as accuracy, speed, or cost. By prioritizing \textbf{accuracy} as a core constraint, ORI ensures that the most reliable model is selected for each subtask or benchmark, while still optimizing for speed and cost where possible. This dynamic selection process allows ORI to maintain high performance under various conditions, seamlessly scaling without sacrificing efficiency or precision.

\end{enumerate}

\section{Related Works}
In the field of Large Language Models (LLMs), efficiently routing tasks to the most suitable model has become a crucial focus of research. The varied capabilities and computational demands of different LLMs mean that no single model uniformly excels at every task or benchmark, highlighting the need for dynamic and adaptive routing systems.

Maurya et al. [10] demonstrated the central challenge of variability in model performance, showing that performance disparities exist between tasks and metrics. In response, researchers have explored strategies for matching tasks to models in real time, balancing performance strengths against operational and economic constraints. A notable example is FrugalGPT [19], proposed by Chen et al., which uses cascading, prompt adaptation, and model approximation to dynamically choose among LLMs such as GPT-4, ChatGPT, and J1-Jumbo. By carefully selecting when to invoke more expensive models, FrugalGPT can reduce operational costs by up to 98\% or achieve higher accuracy for a given budget.

Building on these ideas, the SMOOTHIE method of Guha et al. [12] introduces a label-free routing mechanism using a latent variable graphical model. This approach estimates sample-dependent quality scores for each LLM by comparing model outputs with a latent 'true' output. With no need for labeled data, SMOOTHIE achieves up to a 10-point improvement in routing accuracy over baseline methods. Meanwhile, Jiang et al. [18] contribute LLM-BLENDER, an ensemble framework that strategically combines LLM outputs via pairwise ranking (PAIRRANKER) and generative fusion (GENFUSER). By capitalizing on inter-model variability, LLM-BLENDER consistently outperforms any single model on its own.

Another notable development is the Tryage system [17], which surpasses traditional methods such as Gorilla and GPT3.5 Turbo by using a "perceptive router" to predict downstream performance per request. Tryage even accounts for user constraints, such as model size and recency, through specific flags. This context-sensitive approach proves to be more dynamic and responsive, improving both accuracy and efficiency.

Adding to this evolving landscape, Shnitzer et al. [16] demonstrate how benchmark-driven selection can be treated as a series of binary classification tasks. Their “router” model is trained on performance data from various benchmark sets, which helps it to determine which LLM is best suited for a new task. By leveraging historical performance, Shnitzer et al. show that high-performing models can be chosen more intelligently, thus avoiding the computational overhead of trying multiple LLMs blindly.

Collectively, these studies underscore a rapidly advancing frontier where adaptive and cost-effective routing strategies are integral to LLM deployment. Integrating ideas such as unsupervised routing, ensemble, cascading, and perceptive routers can lead to even more sophisticated systems that unite the goals of accuracy, efficiency, and autonomy. This dynamic interplay of methods points to a promising future for hybrid routing frameworks, where LLMs are selected and combined in real time to maximize performance and minimize cost, ultimately ensuring more reliable, scalable, and economical applications of Large Language Models.

The cumulative insights from these studies illustrate a rapidly evolving landscape where dynamic, cost-effective and intelligent routing systems are integral to the practical deployment of LLMs. These advances set the stage for the development of hybrid approaches that might integrate adaptive, economic and accuracy-focused routing strategies to further improve model selection precision and operational efficiencies. Such integrative efforts could propel the field toward more autonomous and reliable LLM deployments, highlighting a rich avenue for future research.

We introduce ORI (O Routing Intelligence), a routing system that leverages vector space representations and sophisticated categorization algorithms to optimize query-specific performance. ORI is designed to dynamically route queries to the most suitable models, maximizing task-specific accuracy while minimizing computational costs. By intelligently analyzing query characteristics, ORI ensures alignment with benchmark-specific requirements and delivers robust performance across multiple benchmarks. Unlike traditional frameworks, ORI eliminates the dependence on human preference data, reducing potential biases, and enhancing generalization across diverse tasks. Its adaptive architecture allows it to handle varying query complexities efficiently, ensuring consistent performance even in dynamic environments. Using advanced vector space techniques and optimized categorization, ORI minimizes computational overhead, providing a cost-effective solution without compromising performance. Comprehensive evaluations demonstrate the ability of ORI to deliver consistent and high-quality results across various benchmarks, addressing the performance variability observed in other frameworks.

\section{Methodology}
\subsection{Problem Statement}
Single-model strategies often struggle to keep pace with the growing range of tasks and diverse complexities handled by large language models (LLMs). Although existing routing frameworks such as MetaLLM, OptLLM, and RouterDC demonstrate improved performance and cost-effectiveness, they frequently depend on human preference data or exhibit inconsistent results across benchmarks. To address these challenges, we introduce a concise and dynamic system that intelligently routes incoming queries to specialized LLMs. By relying on vector space representations and refined categorization algorithms - rather than human preferences - ORI ensures consistently high accuracy, reduced computational costs, and robust performance across varied benchmarks. This multimodel strategy offers a scalable and efficient alternative to single-model approaches, laying the groundwork for more adaptable and cost-sensitive LLM deployments.

\subsubsection{Definitions \& Notation}
Consider \(P = \{p_1, p_2, \dots, p_n\}\) the set of prompts  to be executed. We have a collection of AI models \(M = \{m_1, m_2, \dots, m_k\}\) available to process these prompts and we use benchmark datasets \(B = \{b_1, b_2, \dots, b_p\}\) to evaluate model performance. When a prompt \(p_i \in P\) is input into model \(m_j \in M\), the model produces a predicted response \(\hat{r}_{i,j} = m_j(p_i)\), which we compare against the ground-truth response \(r_i\). We measure the quality of \(\hat{r}_{i,j}\) with an evaluation function \(\text{Eval}(\hat{r}_{i,j}, r_i)\). Consequently, the \emph{performance function} is:
\[
P(p_i, m_j) \;=\; \text{Eval}\bigl(\hat{r}_{i,j},\, r_i\bigr).
\]
Finally, the \emph{routing function} \(R(p_i, m_j)\) indicates whether a prompt \(p_i\) is assigned to model \(m_j\):
\[
R(p_i, m_j) \;=\;
\begin{cases}
1, & \text{if } p_i \text{ is assigned to } m_j,\\[2pt]
0, & \text{otherwise}.
\end{cases}
\]

\subsubsection{Benchmark Dataset and Model Training}
Each benchmark dataset \(b_k \in B\) is split into a \emph{training set} \(B_{\text{train}} = \{b_{k,\text{train}}\}\) and a \emph{testing set} \(B_{\text{test}} = \{b_{k,\text{test}}\}\). During model training, we construct a K-Nearest Neighbors (KNN) model by transforming the training part of each benchmark into embeddings and clustering them. Thus, the cluster formation and the KNN model itself are built exclusively using the data in \(B_{\text{train}}\).

\subsubsection{Objective Function}
The ORI framework aims to maximize overall query-specific accuracy by routing each prompt to its most suitable model. Formally, it seeks to solve:
\[
\max \;\; \sum_{i=1}^{n} \; P\bigl(p_i, m_j\bigr) \, R\bigl(p_i, m_j\bigr),
\]
where \(P(p_i, m_j)\) measures the accuracy of the model \(m_j\) on prompt \(p_i\), and \(R(p_i, m_j)\) indicates whether \(p_i\) is assigned to \(m_j\).

\subsubsection{Metrics}
Once the models are trained, they are evaluated solely on the test portion \(B_{\text{test}}\). For each model \(m_j\) and prompt \(p_i\) in the test set, we compute the performance function \(P(p_i, m_j)\) by comparing the predicted response \(\hat{r}_{i,j}\) to the actual response \(r_i\). We define each model’s \emph{score} on a particular benchmark \(b_k\) as:
\[
\text{Score}(m_j, b_k)
\;=\;
\frac{\sum_{p_i \in b_{k,\text{test}}} P(p_i, m_j)}{\lvert b_{k,\text{test}} \rvert}
\,\times\, 100.
\]

\subsubsection{Constraint}
To ensure that queries are correctly assigned while striving to maximize accuracy, we impose the following constraint:

\paragraph{Query Assignment Constraint (Each Query to One Model)}
Every prompt \(p_i\) must be assigned to exactly one model:
\[
\sum_{j=1}^{k} R(p_i, m_j) = 1,
\quad \forall p_i \in P.
\]

\subsection{Approach}
\subsubsection{Data Creation and Data Preprocessing}
To build a high-quality training dataset for the ORI model, we followed strict guidelines designed to maintain fairness, broad representativeness, and to prevent data leakage. We began by selecting only the training splits from several prominent benchmark datasets, namely MMLU-PRO, GPQA, MUSR, BBH, IFEVAL and Math Level 5, ensuring that there was no overlap between training and test sets.

From each data set training split, we performed random sampling in a proportionate manner so that all relevant subcategories and question types were evenly represented. This step was critical to avoid an over-representation of samples from any dataset, which could otherwise bias the performance of the model. After sampling, all instances were merged into a single unified training set. During this merging phase, we introduced a new column that labels each instance with its original source dataset, a feature that supports more granular analysis and potential stratified sampling later on.

We also applied a thorough pre-processing pipeline that included handling missing values, correcting data errors, standardizing data formats, and removing outliers or anomalous entries. This ensured that the final dataset was clean, consistent, and well structured for effective model training.

\begin{table}[H]
\centering
\small % Reducing the font size
\begin{tabularx}{\columnwidth}{|l|X|X|}
\hline
\rowcolor{lightgray}
\textbf{Dataset} & \textbf{Description} & \textbf{Key Features} \\ \hline
MMLU-PRO & Massive Multitask Language Understanding benchmark & 16,000 multiple-choice questions across 57 subjects, assessing both world knowledge and problem-solving abilities \\ \hline
GPQA & Graduate-Level Google-Proof QA Benchmark & 448 expert-designed multiple-choice questions in biology, physics, and chemistry, crafted to be challenging for non-experts \\ \hline
MUSR & Multistep Reasoning Narratives & Algorithmically generated complex problems requiring integration of reasoning with long-range context parsing \\ \hline
BBH & BIG-Bench Hard subset & 23 challenging tasks from the BIG-Bench dataset, focusing on multi-step reasoning and language understanding \\ \hline
IFEVAL & Instruction Following Evaluation dataset & 1,054 instruction prompts across 9 categories, designed to test models' adherence to explicit instructions \\ \hline
Math L5 & Advanced Mathematics Benchmark & High-school level competition problems in calculus, linear algebra, and probability, formatted with LaTeX and Asymptote \\ \hline
\end{tabularx}
\caption{ Key Benchmarks Used for Dataset Creation. (MC = Multiple Choice)}
\label{tab:benchmark_datasets}
\end{table}

By carefully using proportionate sampling only from the training splits, we achieved a balanced and diverse dataset that minimizes the risk of data leakage and accurately reflects the domain of each benchmark. This structured process provides a robust foundation for reliable and unbiased model training in the ORI framework.

\subsubsection{Embedding Generator}
We explored various embedding models, including traditional word embeddings (e.g., Word2Vec) and more advanced transformer-based encoders (e.g., BERT). Ultimately, we chose to use the Sentence Transformer due to its proven industry adoption, lightweight architecture, and state-of-the-art performance for semantic similarity tasks. 

Sentence Transformer provides compact vector representations (size 384) that efficiently capture semantic information. This reduced dimensionality, compared to larger transformer models, not only decreases computational overhead, but also preserves the accuracy required for tasks like clustering and classification of queries. By converting each input query into a fixed-size vector, we readily group semantically related queries and enhance downstream performance in tasks such as task classification and query routing. Additionally, the Sentence Transformer framework offers robust pretrained models and fine-tuning options, making it highly adaptable to new domains or specialized requirements in an industrial context.

\subsubsection{Clustering}

We applied both K-Means and Agglomerative Clustering techniques to analyze the structure of our dataset, with the goal of grouping similar queries effectively. For K-Means clustering, we evaluated cluster counts ranging from 2 to 30 using the Silhouette score, which quantifies clustering quality. The Silhouette score \( s(i) \) for a data point \( i \) is calculated as 
\[
s(i) = \frac{b(i) - a(i)}{\max\{a(i),\,b(i)\}},
\]
where \( a(i) \) represents the average distance between the \( i \)-th data point and all other points within the same cluster, while \( b(i) \) denotes the minimum average distance of the \( i \)-th data point to points in any other cluster. Higher silhouette scores indicate better-defined clusters. The results of our K-Means clustering analysis are summarized in Table~\ref{tab:my_table}, where we identified 20 as the optimal number of clusters. However, the corresponding Silhouette scores remained low, with values less than 0.11, suggesting limited cluster separability.

Similarly, we conducted Agglomerative clustering, a hierarchical technique that iteratively merges data points based on their proximity. As with K-Means, we tested cluster counts from 2 to 30 and computed silhouette scores for each configuration. The hierarchical approach showed some improvement in the Silhouette scores as the number of clusters increased, but overall the scores remained relatively low, with the optimal number of clusters also identified as 20 (Table~\ref{tab:my_table}). The Silhouette scores for Agglomerative clustering similarly fell below 0.11, indicating that neither method achieved strong cluster separation.

\begin{figure}[H]
\centering
\includegraphics[width=0.8\textwidth]{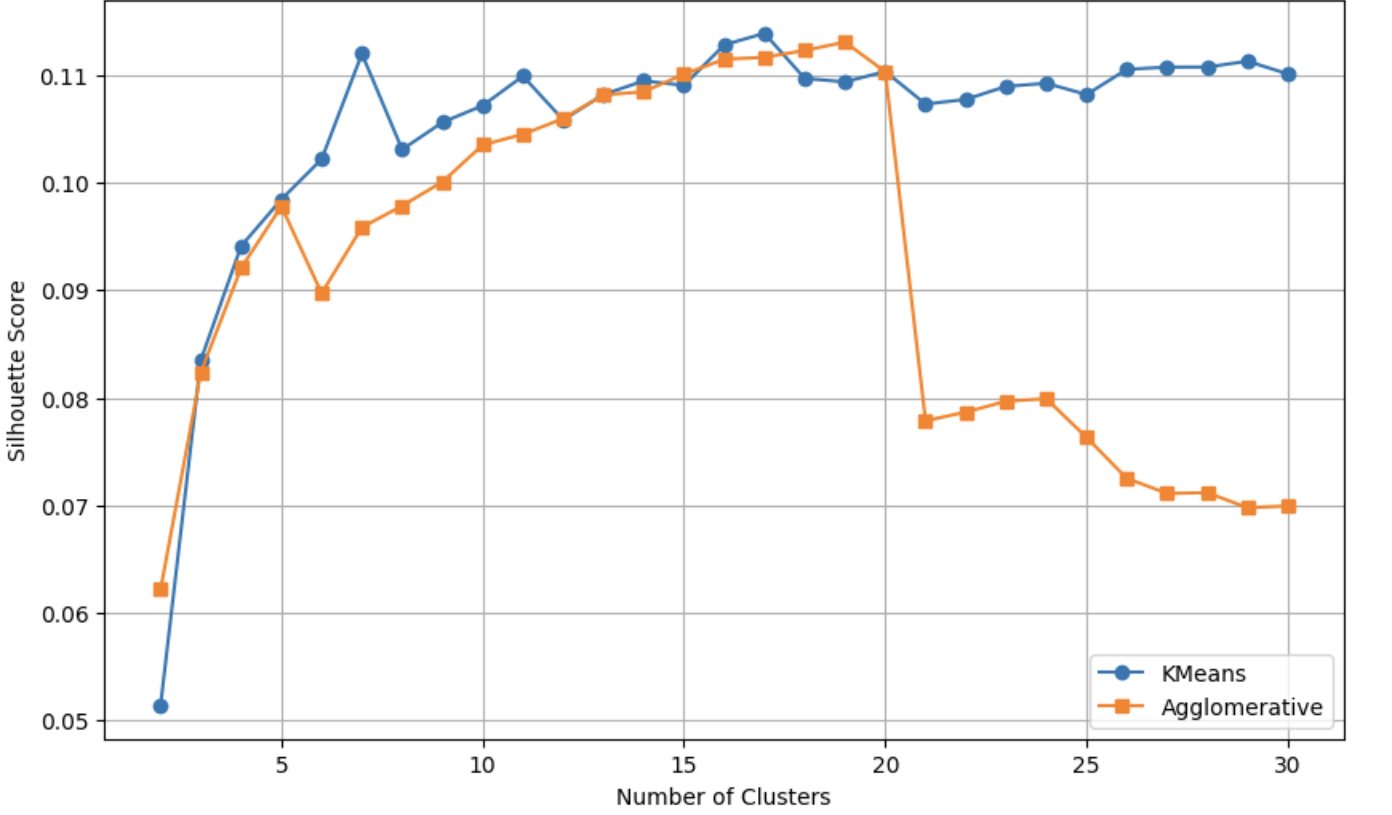} % Replace with your image file path
\caption{Silhouette Score comparison of K-Means and Agglomerative clustering across different cluster counts.}
\label{fig:example_image}
\end{figure}

A visual comparison of the Silhouette scores for K-Means and Agglomerative clustering across different cluster counts is presented in Figure~\ref{fig:example_image}. This figure highlights that while both methods exhibited similar trends, neither significantly outperformed the other in terms of cluster quality, as indicated by consistently low Silhouette scores. These results suggest that the inherent structure of the data may not lend itself well to distinct clustering or that alternative clustering techniques may be required for better performance.

\begin{table}[ht]
\centering
\begin{tabular}{|l|l|l|}
\hline
\rowcolor{lightgray}
\textbf{Clustering Method} & \textbf{Optimal Clusters} & \textbf{Silhouette Score Range} \\
\hline
K-Means & 20 & Low (< 0.11) \\
\hline
Agglomerative & 20 & Low (< 0.11) \\
\hline
\end{tabular}
\caption{Optimal cluster count and corresponding Silhouette score range for K-Means and Agglomerative clustering.}
\label{tab:my_table}
\end{table}

\textbf{Cluster Analysis}:
The clusters were mapped to the benchmarks by grouping the data points according to cluster labels. The distribution of benchmarks within each cluster was analyzed to identify dominant benchmarks. This step helped to route queries more effectively by associating clusters with specific benchmarks.

\begin{enumerate}
    \item \textbf{K-Means Cluster Assignment}:

Let \( \{ p_1, p_2, \ldots, p_n \} \subset \mathbb{R}^d \) represent a set of \( n \) prompts embedded in a \( d \)-dimensional space. The goal is to assign each prompt to one of \( K \) clusters based on its proximity to the cluster centroids. The assignment is determined by:

\[
c(p_i) = \arg\min_{k \in \{1, \ldots, K\}} \bigl\| p_i - \boldsymbol{\mu}_k \bigr\|^2,
\]

where \( c(p_i) \) denotes the cluster label assigned to prompt \( p_i \), and \( \boldsymbol{\mu}_k \) is the centroid of the \( k \)-th cluster. Each prompt is thus assigned to the cluster whose centroid is nearest in terms of Euclidean distance, ensuring that prompts with similar semantic meanings are grouped together.
 \item \textbf{Benchmark Distribution in Clusters}: 
The dataset is grouped by cluster labels, with each prompt assigned to a cluster based on the K-Means algorithm. This grouping helps analyze how the prompts within each cluster are distributed across benchmarks.
Assume that we have \( B \) benchmarks, denoted \( b_1, b_2, \ldots, b_p \). After clustering, define:
\[
D_{k,j} = \#\{p_i \mid c(p_i) = k,\; p_i \in b_j \},
\]
where \( D_{k,j} \) is the count of prompts in cluster \( k \) belonging to benchmark \( b_j \).

We convert counts to proportions for each cluster:
\[
p_{k,j} = \frac{D_{k,j}}{\sum_{\ell=1}^{p} D_{k,\ell}}.
\]
This \( p_{k,j} \) indicates the fraction of prompts in cluster \( k \) that come from benchmark \( b_j \).
For instance, if Cluster 1 contains primarily MMLU-PRO and BBH benchmarks, it suggests that this cluster specializes in complex reasoning and multistep queries.
\item \textbf{Cluster-Benchmark Mapping:} Clusters were analyzed for dominance of the benchmark, showing clear associations between specific clusters and benchmarks. For example:

To identify the benchmark that dominates cluster \( k \), we select the index \( j^*(k) \) that satisfies:
\[
j^{*}(k) = \arg\max_{1 \le j \le p} D_{k,j}.
\]

Hence, \( b_{j^*(k)} \) is the most frequent (or dominant) benchmark in cluster \( k \), where:

\begin{itemize}
    \item \( D_{k,j} \) is the count of prompts in cluster \( k \) belonging to benchmark \( b_j \).
    \item \( p \) represents the total number of benchmarks.
    \item \( b_{j^*(k)} \) is the dominant benchmark associated with cluster \( k \).
\end{itemize}

\item \textbf{Benchmark Mapping by Cluster}:

This detailed mapping aids in identifying which benchmarks dominate each cluster, ensuring appropriate routing, and enhancing query specialization.
\begin{enumerate}
    \item \textbf{Silhouette Scores}: 
    \begin{itemize}
        \item The scores showed variability, but generally improved with more clusters.
        \item The scores ranged from 0.08 to 0.11, indicating moderate cluster quality.
    \end{itemize}
    \item \textbf{Optimal Clusters}: 
    \begin{itemize}
        \item K-Means identified an optimal number of clusters around 20, with diminishing improvements beyond this point.
        \item Agglomerative clustering showed comparable results but lower overall scores.

    \end{itemize}
\item \textbf{Embedding Visualization with t-SNE}:
    To better understand the clustering of prompts, we employed \textit{ t-distributed stochastic neighbor embedding (t-SNE)} to visualize high-dimensional embeddings in a two-dimensional space.
    These visual tools provided a deeper understanding of the clustering structure and the alignment of new prompts with existing benchmarks, enhancing insights into the ORI system's routing strategy.

\begin{figure}[H]
    \centering
    \includegraphics[width=0.9\linewidth]{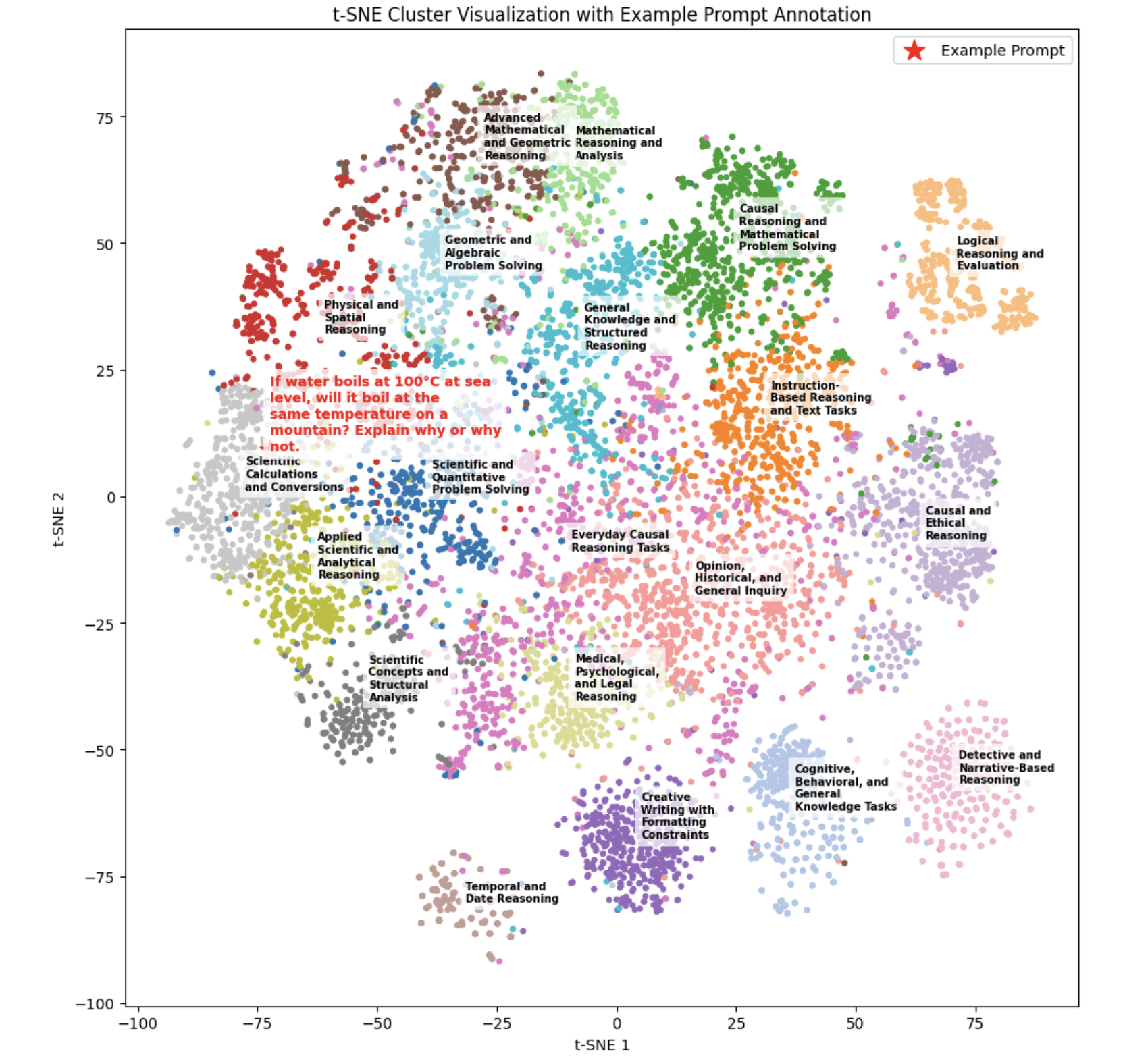}
    \caption{t-SNE Clusters and Example Prompt Annotation. The red star indicates the new prompt, pointing from its predicted cluster centroid. Each cluster is labeled based on its dominant benchmark and analyzed thematic content.}
    \label{fig:tsne_clusters}
\end{figure}

In this visualization, benchmarks are broken down into finer categories such as Scientific and Quantitative Problem Solving, Causal Reasoning and Mathematical Problem Solving, Detective and Narrative-Based Reasoning, and Temporal and Date Reasoning, among others. This granular clustering allows for identifying task-specific performance nuances that may be masked at the broader benchmark level. For example, while a model may perform well on the general Scientific Concepts and Structural Analysis benchmark, it might struggle with sub-tasks under Applied Scientific and Analytical Reasoning or Scientific Calculations and Conversions.

\item \textbf{Model Mapping to Benchmark}: 
To ensure optimal performance, models were selected from the Hugging Face leaderboard based on their ability to outperform others in their respective benchmarks. For each benchmark, only the top-performing models—those achieving the highest scores relative to competitors—were mapped. This approach guarantees that each task or sub-task identified in the clustering analysis is paired with the most capable model, enhancing overall system accuracy and efficiency. 

The table below summarizes the benchmarks alongside their corresponding best-performing models and scores:

\begin{table}[H]
\centering
\scriptsize 
\begin{tabular}{|l|l|l|c|}
\hline
\rowcolor{lightgray}
\textbf{Benchmark} & \textbf{Name} & \textbf{Model} & \textbf{Score} \\ \hline
MMLU      & Massive Multitask Language Understanding & Qwen2.5-72B          & 82.3 \\ \hline
GPQA      & Google-Proof Q\&A                        & Qwen2.5-72B          & 49.0 \\ \hline
Math-L5   & Math Level 5                             & Qwen2.5-72B          & 52.7 \\ \hline
MuSR      & Multistep Reasoning                      & Calme-2.4-78B        & 36.37 \\ \hline
BBH       & BIG-Bench Hard                           & Deepseek-67B         & 78.8 \\ \hline
IFEval    & Instruction Eval                         & Llama-3.3-70B        & 92.1 \\ \hline
\end{tabular}
\caption{Benchmark performance of top models selected from the Hugging Face leaderboard. Models: meta-llama/Meta-Llama-3.3-70B (Llama-3.3-70B), MaziyarPanahi/calme-2.4-rys-78b (Calme-2.4-78B), deepseek-ai/deepseek-llm-67b-chat (Deepseek-67B), Qwen/Qwen2.5-72B (Qwen2.5-72B).}
\label{tab:benchmark_models}
\end{table}

This model-to-benchmark mapping complements the clustering analysis presented earlier. As illustrated in the t-SNE visualization , distinct clusters correspond to specific reasoning or task categories. By associating these clusters with the best-performing models for the relevant benchmarks, we ensure that even granular sub-tasks are handled by models that excel in their domain. This method mitigates the risk of performance degradation on sub-tasks that may otherwise be overshadowed in broad benchmark evaluations, enabling more precise and context-sensitive model selection.

\end{enumerate}
\end{enumerate}
\subsubsection{Routing Process}
When a new prompt \(p_i\) arrives, it is converted into an embedding:
\[
x_i \;=\; \text{Embed}\bigl(p_i\bigr).
\]
Using the trained cluster model, we identify its nearest cluster:
\[
c\bigl(p_i\bigr) \;=\; \arg\min_{k} \,\|x_i - \mu_k\|^2,
\]
where \(\mu_k\) is the centroid for cluster \(k\). To determine which model handles \(p_i\), we use \emph{benchmark dominance} within the identified cluster, routing \(p_i\) to the model that achieves the highest score on the dominant benchmark of the cluster:
\[
R(p_i, m_j) \;=\; 1
\quad \text{if} \quad
m_j \;=\; \arg\max_{m_j} \,\text{Score}\bigl(m_j, b_{\text{dominant}}\bigr).
\]

\section{Results}
\subsection{ORI Benchmarking}

The ORI model demonstrated effective routing and query-specific performance improvements on standard benchmarks. The performance comparison between ORI and other large language models is summarized in Table 5. ORI outperformed competitors in most benchmarks, showcasing its superiority in query-specific optimization.

\begin{itemize}
    \item BBH Benchmark: ORI achieved a score of 78.0, which is competitive with top-performing models such as Deepseek-67B (78.8).
    \item MMLU Benchmark: ORI outperformed all models, achieving the highest score of 85.0, compared to the next best, Samba-CoE v0.3, with 83.7.
    \item MuSR Benchmark: ORI achieved a score of 38.1, which is better than most routing models and on par with large models like Calme-2.4-78B (36.3).
    \item ARC Benchmark: ORI matched the best performing models (Meta-Llama-3-70B) with a perfect score of 93.0.
\end{itemize}

\begin{table}[H]
\centering
\small
\begin{tabular}{|l|c|c|c|c|}
\hline
\rowcolor{lightgray}
\textbf{Model} & \textbf{BBH} & \textbf{MMLU} & \textbf{MuSR} & \textbf{ARC} \\
\hline
\multicolumn{5}{|c|}{\textit{Large Models (>60B)}} \\
\hline
Llama-3-70B        & 78.0      & 82.1         & 21.4    & \textbf{93.0}  \\ \hline
Qwen2-72B          & 78.0      & 82.3         & 19.7  & 69.0  \\ \hline
Deepseek-67B       & \textbf{78.8}      & 62.8        & 23.9  & 90.0  \\ \hline
Calme-2.4-78B      & 62.16     & 66.6      & 36.3  & -   \\ \hline
\multicolumn{5}{|c|}{\textit{Routing Models}} \\
\hline
Route LLM          & -         & 56.8          & -      & -     \\ \hline
Samba-CoE v0.3     & -         & 83.7          & -      & 72.2  \\ \hline
Samba-CoE v0.2     & -         & 63.3          & -      & 72.8  \\ \hline
RouterDC           & -         & 61.1          & -      & 58.5  \\ \hline
\multicolumn{5}{|c|}{\textit{Smaller Models (<60B)}} \\
\hline
Gemma-7B           & -         & 64.6          & -      & 61.1  \\ \hline
Mixtral-8x7B       & -         & 71.2          & -      & 70.2  \\ \hline
Mistral-7B         & -         & 62.1          & -      & 49.4  \\ \hline
\multicolumn{5}{|c|}{\textit{Our Models}} \\
\hline
\textbf{ORI}       & 78.0      & \textbf{85.0} & \textbf{38.1} & \textbf{93.0} \\ \hline
\end{tabular}
\caption{Model Performance Comparison. Best scores in bold.}
\label{tab:model_comparison}
\end{table}

\subsection{Cost and Performance Analysis}

This section evaluates the efficiency and cost-effectiveness of the ORI framework, focusing on token generation speed, response times, and overall cost. The analysis utilizes the Massive Multitask Language Understanding (MMLU) benchmark to assess performance across various models.

In this analysis, the MMLU test was used, consisting of 13,869 prompts.

\subsubsection{Cost Analysis}

The cost analysis evaluates the total financial expenditure required to process the dataset using different models. The results indicate that ORI offers balanced cost-efficiency, with a total cost of \$1.94, which is slightly higher than Meta-Llama-3-70B (\$1.49) but lower than Deepseek-LLM-67B (\$7.06) and Calme-2.4 (\$2.24).

Although Meta-Llama-3-70B appears more cost-effective in isolation, ORI's ability to optimize both speed and latency ensures that the slight increase in cost translates into broader efficiency gains. Models like Deepseek-LLM-67B incur significantly higher costs without proportional improvements in speed or latency, making ORI a cost-effective choice for scenarios where both performance and budget are critical.

\begin{table}[H]
\centering
\small
\begin{tabular}{|l|c|}
\hline
\rowcolor{lightgray}
\textbf{Model} & \textbf{Total Cost (USD)} \\ \hline
Meta-Llama-3-70B        & \textbf{1.49} \\ \hline
Qwen2-72B               & 2.35 \\ \hline
Deepseek-LLM-67B        & 7.06 \\ \hline
Calme-2.4               & 2.24 \\ \hline
\textbf{ORI}            & 1.94 \\ \hline
\end{tabular}
\caption{Total cost comparison across models using the MMLU dataset. ORI demonstrates competitive cost-efficiency relative to other models.\protect\footnotemark}
\label{tab:cost_comparison}
\end{table}
\footnotetext{Model and pricing information provided by \url{https://aimlapi.com}; the models were accessed via their API for use in the routing model. Only one provider was used for the experiment to ensure an unbiased analysis.}

\subsubsection{Token Generation Speed}

Token generation speed reflects how quickly each model produces outputs. ORI achieved a speed of 375 tokens/second, outperforming Qwen2-72B (131 tokens/second) and Calme-2.4 (128 tokens/second), while surpassing Deepseek-LLM-67B (60 tokens/second). Although Meta-Llama-3-70B leads with 444 tokens/second, the gap is marginal when considering ORI's superior cost-efficiency and competitive latency.

The strong token generation speed of ORI is a result of its intelligent query routing, which assigns tasks to the most suitable models based on complexity. This ensures efficient processing of both simple and complex queries, allowing ORI to achieve high throughput without compromising accuracy. While Meta-Llama-3-70B is faster, ORI's combination of speed, cost, and latency makes it a more versatile option.

\begin{table}[H]
\centering
\small
\begin{tabular}{|l|c|}
\hline
\rowcolor{lightgray}
\textbf{Model} & \textbf{Tokens/Second} \\ \hline
Meta-Llama-3-70B        & \textbf{444} \\ \hline
Qwen2-72B               & 131 \\ \hline
Deepseek-LLM-67B        & 60 \\ \hline
Calme-2.4               & 128 \\ \hline
\textbf{ORI}            & 375 \\ \hline
\end{tabular}
\caption{Token generation speed comparison across models using the MMLU dataset. ORI balances speed and cost efficiency, outperforming several models.\protect\footnotemark}
\label{tab:speed_comparison}
\end{table}
\footnotetext{Token generation speeds were calculated by dividing the total number of tokens generated by the total processing time (in seconds).}

\subsubsection{Latency Analysis}
Latency measures the total time taken to process the dataset, reflecting the ability of the model to deliver timely responses. ORI achieved a total latency of 562 seconds, which is significantly better than Qwen2-72B (605 seconds) and Calme-2.4 (608 seconds). 
The efficient latency of ORI is a result of its dynamic routing system, which directs tasks to the most suitable models, minimizing bottlenecks. Although Meta-Llama-3-70B offers slightly faster processing, ORI's balanced performance across all key metrics makes it a more reliable and cost-effective solution for varied tasks.

\begin{table}[H]
\centering
\small
\begin{tabular}{|l|c|}
\hline
\rowcolor{lightgray}
\textbf{Model} & \textbf{Total Latency (seconds)} \\ \hline
Meta-Llama-3-70B        & \textbf{531} \\ \hline
Qwen2-72B               & 605 \\ \hline
Deepseek-LLM-67B        & 731 \\ \hline
Calme-2.4               & 608 \\ \hline
\textbf{ORI}            & 562 \\ \hline
\end{tabular}
\caption{Latency comparison across models using the MMLU dataset. ORI achieves efficient response times while maintaining cost-effectiveness.}
\label{tab:latency_comparison}
\end{table}

\subsubsection{Comprehensive Performance Evaluation}

When cost, speed, and latency are collectively evaluated, ORI emerges as a well-balanced and efficient solution. Although Meta-Llama-3-70B has the lowest cost and fastest token generation speed, ORI's slight trade-offs in these areas are justified by its competitive latency and overall cost-efficiency. Models like Deepseek-LLM-67B and Calme-2.4 exhibit higher costs and slower speeds, without providing proportional benefits in latency, reducing their overall value.

In conclusion, ORI offers a compelling combination of cost-efficiency, fast token generation, and low latency. Its ability to adaptively route queries ensures that it consistently delivers strong performance across various metrics, making it an ideal choice for real-world applications that demand a balance between performance and operational costs.

\subsection{Key Observations}
\begin{enumerate}
\item \textbf{Cost Efficiency}: ORI offers a competitive cost structure, being more affordable than most models while slightly more expensive than Meta-Llama-3-70B.
\item \textbf{Token Generation Speed}: ORI achieves high token generation speed, outperforming several models, and providing quick response times.
\item \textbf{Latency Performance}: ORI maintains efficient response times, demonstrating lower latency compared to many models, ensuring rapid query processing.
\end{enumerate}

\section{Conclusion}

The ORI (O Routing Intelligence) framework presents a novel approach to query-specific performance optimization and cost-efficient deployment of large language models (LLMs). By integrating advanced clustering techniques, embedding-based representations, and benchmark mapping, ORI effectively addresses the challenges of routing queries to the most appropriate models. This ensures a balance between computational efficiency and query-specific accuracy, a critical need in modern NLP applications.

ORI achieves state-of-the-art performance on standard benchmarks, recording an impressive 85\% on MMLU, 78\% on BBH, 38.1\% on MuSR, and 93\% on ARC. These results underscore its ability to match or exceed the performance of top-tier models such as Meta-Llama-3-70B and Deepseek-67B. The framework's ability to leverage dynamic query routing ensures that the query is processed by models with optimal capabilities, thereby enhancing query-specific outcomes.

In addition to its performance advantages, ORI demonstrates significant potential for resource efficiency. Its integration of balanced sampling, clustering, and embedding-based visualizations ensures fair data set representation, reducing biases while maintaining consistent results. These optimizations position ORI as a practical and scalable solution for deploying LLMs in diverse operational contexts.

Looking ahead, ORI sets the foundation for further exploration of cost metrics, latency optimizations, and extensions to more complex, interdisciplinary queries. Its adaptive and efficient design offers a robust way forward for intelligent routing in NLP systems, bridging the gap between high performance and computational sustainability. By addressing the dual challenges of accuracy and resource utilization, ORI paves the way for more accessible and scalable LLM applications.

\section{References}
\begin{enumerate} 
    \item Brown, T. B., Mann, B., Ryder, N., Subbiah, M., Kaplan, J., Dhariwal, P., ... \& Amodei, D. (2020). \textbf{Language Models are Few-Shot Learners}. arXiv. \url{https://arxiv.org/abs/2005.14165}
    \item Chowdhery, A., Narang, S., Devlin, J., Bosma, M., Mishra, G., Roberts, A., ... \& Petrov, S. (2022). \textbf{PaLM: Scaling Language Modeling with Pathways}. arXiv. \url{https://arxiv.org/abs/2204.02311}
    \item Vaswani, A., Shazeer, N., Parmar, N., Uszkoreit, J., Jones, L., Gomez, A. N., Kaiser, Ł., \& Polosukhin, I. (2017). \textbf{Attention Is All You Need}. arXiv. \url{https://arxiv.org/abs/1706.03762}
    \item Radford, A., Narasimhan, K., Salimans, T., \& Sutskever, I. (2018). \textbf{Improving Language Understanding by Generative Pre-Training}. arXiv. \url{https://arxiv.org/abs/1801.00173}
    \item Nguyen, Q. H., Hoang, D. C., Decugis, J., Manchanda, S., Chawla, N. V., \& Doan, K. D. (2024). \textbf{MetaLLM: A High-performant and Cost-efficient Dynamic Framework for Wrapping LLMs}. arXiv. \url{https://arxiv.org/abs/2407.10834}
    \item Liu, Y., Zhang, H., Miao, Y., Le, V.-H., \& Li, Z. (2024). \textbf{OptLLM: Optimal Assignment of Queries to Large Language Models}. arXiv. \url{https://arxiv.org/abs/2405.15130}
    \item Chen, S., Jiang, W., Lin, B., Kwok, J. T., \& Zhang, Y. (2024). \textbf{RouterDC: Query-Based Router by Dual Contrastive Learning for Assembling Large Language Models}. arXiv. \url{https://arxiv.org/abs/2409.19886}
    \item Ong, I., Almahairi, A., Wu, V., Chiang, W.-L., Wu, T., Gonzalez, J. E., Kadous, M. W., \& Stoica, I. (2024). \textbf{RouteLLM: Learning to Route LLMs with Preference Data}. arXiv. \url{https://arxiv.org/abs/2406.18665}
    \item Aggarwal, P., Madaan, A., Anand, A., Potharaju, S. P., Mishra, S., Zhou, P., Gupta, A., Rajagopal, D., Kappaganthu, K., Yang, Y., Upadhyay, S., Faruqui, M., \& Mausam. (2024). \textbf{AutoMix: Automatically Mixing Language Models}. arXiv. \url{https://arxiv.org/abs/2310.12963v4}
    \item Maurya, K. K., Srivatsa, K. V. A., \& Kochmar, E. (2024). \textbf{SelectLLM: Query-Aware Efficient Selection Algorithm for Large Language Models}. arXiv. \url{https://arxiv.org/abs/2408.08545}
    \item Zhuang, R., Wu, T., Wen, Z., Li, A., Jiao, J., \& Ramchandran, K. (2024). \textbf{EmbedLLM: Learning Compact Representations of Large Language Models}. arXiv. \url{https://arxiv.org/abs/2410.02223v2}
    \item Guha, N., Chen, M. F., Chow, T., Khare, I. S., \& Ré, C. (2024, December 9). \textbf{SMOOTHIE: Label-Free Language Model Routing}. arXiv. \url{https://arxiv.org/abs/2412.04692v1}
    \item Feng, T., Shen, Y., \& You, J. (2024). \textbf{GraphRouter: A Graph-Based Router for LLM Selections}. arXiv. \url{https://arxiv.org/abs/2410.03834}
    \item Li, Z., \& Zhou, T. (2024). \textbf{Your Mixture-of-Experts LLM is Secretly an Embedding Model for Free}. arXiv. \url{https://arxiv.org/abs/2410.10814v2}
    \item Zhang, J., Krishna, R., Awadallah, A. H., \& Wang, C. (2023). \textbf{EcoAssistant: Using LLM Assistant More Affordably and Accurately}. arXiv. \url{https://arxiv.org/abs/2310.03046}
    \item Shnitzer, T., Ou, A., Silva, M., Soule, K., Sun, Y., Solomon, J., Thompson, N., \& Yurochkin, M. (2023). \textbf{Large Language Model Routing with Benchmark Datasets}. arXiv. \url{https://arxiv.org/abs/2309.15789}
    \item Hari, S. N., \& Thomson, M. (2023). \textbf{Tryage: Real-time, Intelligent Routing of User Prompts to Large Language Models}. arXiv. \url{https://arxiv.org/abs/2308.11601}
    \item Jiang, D., Ren, X., \& Lin, B. Y. (2023). \textbf{LLM-BLENDER: Ensembling Large Language Models with Pairwise Ranking and Generative Fusion}. arXiv. \url{https://arxiv.org/abs/2306.02561v3}
    \item Chen, L., Zaharia, M., \& Zou, J. (2023). \textbf{FrugalGPT: How to Use Large Language Models While Reducing Cost and Improving Performance}. arXiv. \url{https://arxiv.org/abs/2305.05176}
\end{enumerate}
\end{document}